\documentclass[letterpaper, 10pt, conference]{ieeeconf}  
\renewcommand{\thefigure}{\arabic{figure}} 

\IEEEoverridecommandlockouts                              

\usepackage{amsmath} 
\usepackage{amssymb}  
\usepackage{url}
\usepackage{subfigure} 
\usepackage{graphicx}
\usepackage{subcaption}
\usepackage{booktabs} 
\usepackage{array} 
\title{\LARGE \bf
Abdominal Undulation with Compliant Mechanism Improves Flight Performance of Biomimetic Robotic Butterfly
}

\begin{document}


\author{
    Xuyi Lian$^{1, \dagger}$, Mingyu Luo$^{2, \dagger}$, Te Lin$^{2}$, Chen Qian$^{1, \star}$ and Tiefeng Li$^{1, \star}$%
    \thanks{This work was supported by China Postdoctoral Science Foundation Funded Project under Grant 2022M722912, National Natural Science Foundation of China(T2125009), "Pioneer" R\&D Program of Zhejiang (Grant no. 2023C03007).}
    \thanks{${\dagger}$These authors contributed equally to this work.}
    \thanks{${\star}$These authors are corresponding authors.}
    \thanks{$^{1}$These authors are with School of Aeronautics and Astronautics, Zhejiang University, Zhejiang, China.
    {\tt\small liam.lianxy@zju.edu.cn}, {\tt\small chainplain1028@gmail.com}, {\tt\small litiefeng@zju.edu.cn}}
   \thanks{$^{2}$These authors are with School of Mechanical Engineering, Zhejiang University, Zhejiang, China.
    {\tt\small luomy@zju.edu.cn},
    {\tt\small 1209348538@qq.com}}
}

\maketitle
\thispagestyle{empty}
\pagestyle{empty}

\begin{abstract}
This paper presents the design, modeling, and experimental validation of a biomimetic robotic butterfly (BRB) that integrates a compliant mechanism to achieve coupled wing-abdomen motion. Drawing inspiration from the natural flight dynamics of butterflies, a theoretical model is developed to investigate the impact of abdominal undulation on flight performance. To validate the model, motion capture experiments are conducted on three configurations: a BRB without an abdomen, with a fixed abdomen, and with an undulating abdomen. The results demonstrate that abdominal undulation enhances lift generation, extends flight duration, and stabilizes pitch oscillations, thereby improving overall flight performance. These findings underscore the significance of wing-abdomen interaction in flapping-wing aerial vehicles (FWAVs) and lay the groundwork for future advancements in energy-efficient biomimetic flight designs.

\end{abstract}

\section{Introduction}

Flapping-wing aerial vehicles (FWAVs) have demonstrated advantages in maneuverability, energy efficiency, and adaptability, making them ideal for potential applications such as aerial surveillance, environmental monitoring, as well as search-and-rescue missions. 
Over past decades, significant progress has been made in designing bio-inspired FWAVs and exploring underlying biological mechanisms, including mimicking of hummingbirds \cite{Tu-2020}, \cite{Fei-2023}, bees \cite{Wood-2013}, dragonflies \cite{Dileo-2009}, bats \cite{Ramezani-2016}, \cite{Hoff-2021}, \cite{Fan-2021} and birds \cite{Huang-2022}.

In contrast to these species, butterflies exhibit an expansive wing area and increased flexibility, and they display a lower wingbeat frequency of approximately 10 Hz, which is lower than that of other insects, typically ranging from 25 to 40 Hz \cite{Zhang-2024}. 
Therefore, butterfly flight mechanisms are more unique, with more complex flight behaviors and aerodynamic characteristics, making the development of butterfly-inspired FWAVs more challenging. 
It is reported that butterflies possess relatively broad forewings and hindwings, with the wings on the same side beating in near synchrony and exhibiting a large flapping amplitude \cite{Fujikawa-2008}.

Butterflies adjust their body posture, especially abdominal movements, for flight stability.
Previous studies emphasize the importance of abdominal movements in insect flight performance \cite{Zanker-1988, Baader-1990, Hedrick-2006}. 
Recently, increasing attention has been given to observing abdominal movement in butterflies. 

\begin{figure}[t]
    \centering
    \includegraphics[width=0.75\linewidth]{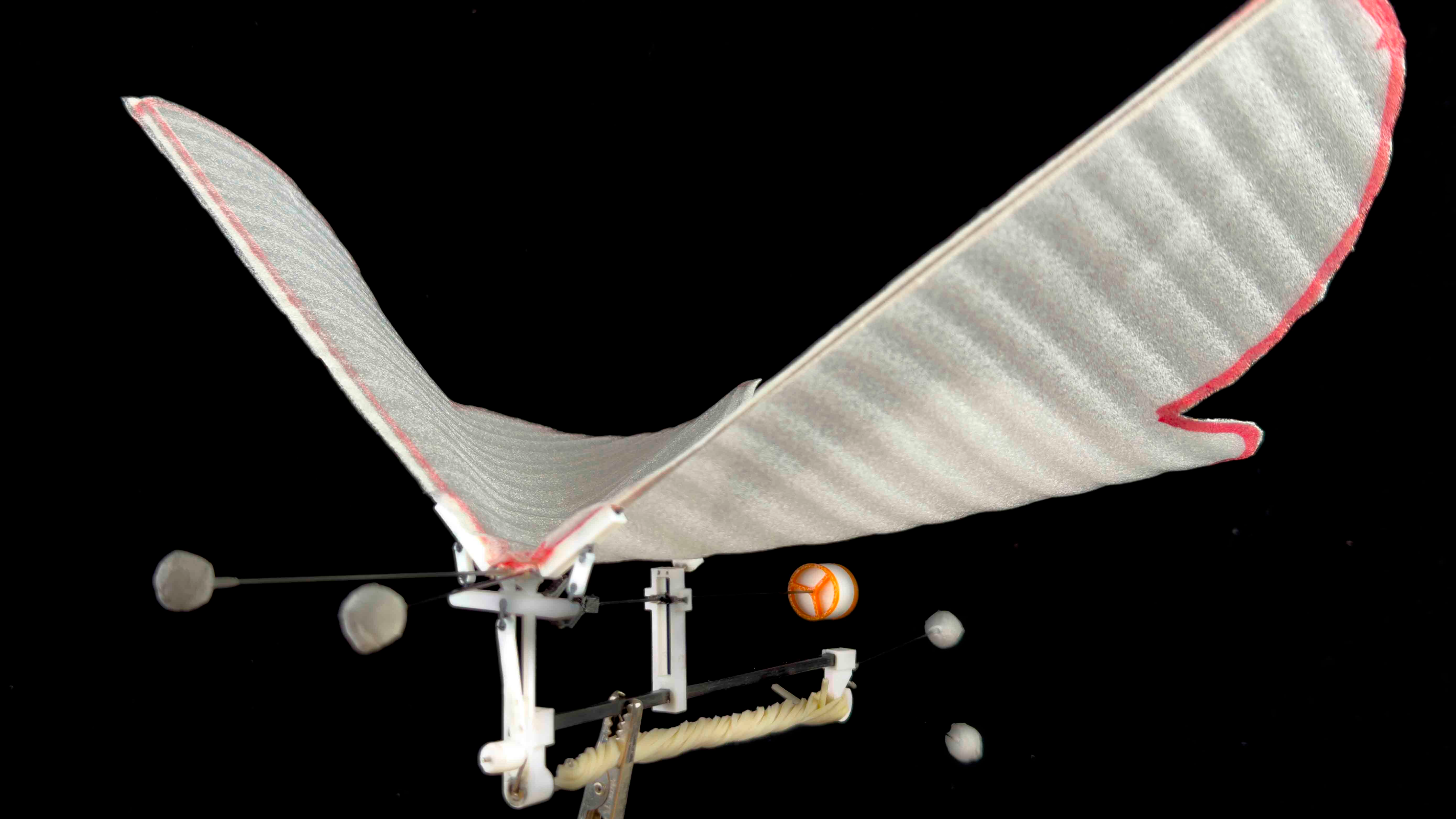} 
    \caption{Illustration of the proposed biomimetic robotic butterfly (BRB) with compliant mechanism.}
    \label{fig:prototype}
\end{figure}

Because the butterfly wings attached to the thorax have a relatively high moment of inertia, aerodynamic and inertial forces cause the thorax to pitch in sync with the wingbeats.
Under various flight conditions, like takeoff, hovering, and forward flight, the abdomen swings in response to these thoracic oscillations \cite{Fujikawa-2008, Senda-2012-1, Zhang-2021}. 
Consequently, it attracts more and more attentions to the study of the abdominal movement effects for butterfly flight. Jayakumar et al. \cite{Jayakumar-2018} numerically analyzed the impact of abdominal movement on pitch stability using a simplified 2D butterfly model,
which considers the inertia effects from the flapping wings. 
It is then demonstrated that abdominal motion significantly enhances short-period flight stability. 
Similarly, Tejaswi et al. \cite{Tejaswi-2021} find that abdominal undulation in monarch butterflies reduces energy consumption by 10.7\% during hovering and 6.1\% during forward-climbing flight, while simultaneously improving stability by increasing the convergence rate of the slowest flight modes.

To date, FWAVs modeled after butterflies remain uncommon. 
Festo Corp. developed the eMotion-Butterfly, featuring independently motor-driven dual wings with integrated forewings and hindwings, weighing about 38 g, capable of level flight, climbing, and turning \cite{Festo}.
Additionally, Huang et al. \cite{Huang-2024} developed a servo-driven, biomimetic robotic butterfly with independent wing control, finding that despite its larger size and higher Reynolds number compared to its natural counterpart, the robot can still exhibit butterfly-like wing-body coupling during climbing flight. 
Nevertheless, these biomimetic butterfly models neglect the important role of abdominal swinging to reduce the pitch oscillation during large-inertia wing flapping motion. 
\begin{figure*}[t]
    \centering
    \includegraphics[width=0.9\textwidth]{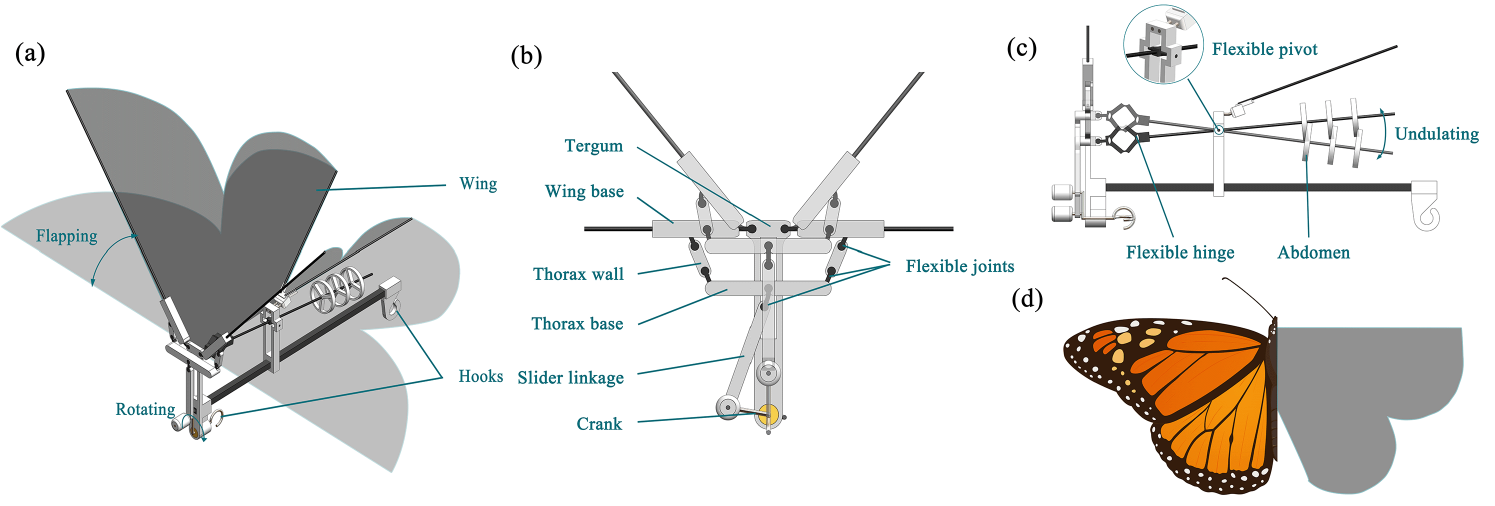} 
    \caption{(a) The prototype robot showing degree of freedom of flapping wing. (b) Flapping mechanism, (c) abdominal undulation mechanism and (d) wing design based on morphological similarity.}
    \label{fig:3d}
\end{figure*}
The major engineering challenge in developing FWAVs with wing-abdomen coupling mechanism stems from the reliance on rigid mechanisms, exemplified by  hinges, axles, and bearings, which increase complexity, weight, and inconvenience in finding center-of-gravity configuration. 
In contrast, biological systems achieve motion through flexible structures rather than rigid connections, which is ubiquitously shown in hearts, elephant trunks, and bee wings \cite{Howell-2013}.

Compliant mechanisms, generating movement through elastic deformation, closely mimic biological structures and offer effective solutions to FWAV design challenges. 
Wood developed a 60 mg FWAV with a 3 cm wingspan, using carbon fiber reinforced polymer for the exoskeleton and polyimide film for flexible hinges, integrating compliant thoracic structures and elastic energy storage in piezoelectric actuators to enable efficient vertical takeoff \cite{Wood-2007}.
Moreover, Chin et al. designed an insect-inspired compliant thoracic mechanism, achieving power savings of up to 31\% compared to rigid-body designs \cite{Chin-2014}.
Although the application of compliant mechanism in FWAV design remains limited, its potential implementation is exceptionally promising due to the friction-less nature and the ability to store elastic energy.

In this work, we design a  biomimetic robotic butterfly (BRB) with compliant mechanisms to achieve coupled wing-abdomen motions, demonstrating its positive effects on flight efficiency, which is clearly shown in Fig.\ref{fig:prototype}. 
We also develop a theoretical model to analyze wing flapping and abdominal undulation, investigating their effects on flapping patterns and average lift. 
Motion capture experiments are conducted to collect flight trajectories and attitude performances, where cases with and without abdominal undulation are further compared. 
The results validate the role of abdominal motion in enhancing flight performance. 
The main contributions of this are summarized as follows: 
\begin{enumerate}
    \item A butterfly-inspired FWAV using a compliant mechanism to achieve coupled wing-abdomen motion is successfully developed. 
    \item Through dynamic modeling and experimental analysis, the flight performance improvement from abdominal undulation is validated.
\end{enumerate}

The remainder of this paper is structured as follows. Section II details the design and fabrication of the proposed prototype. After that, the theoretical analysis of the bio-inspired flapping and abdominal undulation mechanisms are perfromed in Section III. Section IV describes the flight tests of the BRB and analyzes the results. Finally, Section V concludes the study.

\section{Design and Fabrication}

\subsection{Flapping Mechanism}

The flapping mechanism of butterflies is driven by indirect muscle contractions. Butterfly wings evolved from the exoskeleton, with their ends hinged to the tergum, and an additional hinge connecting the wings to the thoracic wall. Within the thorax, alternating contractions of longitudinal muscles (connecting the front and rear) and dorsoventral muscles (connecting the top and bottom) raise and lower the tergum. According to the principle of leverage, these movements result in the upward and downward flapping of the wings. Consequently, this flapping mechanism can be regarded as a single-degree-of-freedom (DOF) vibratory system \cite{Senda-2012-1}.
 
Based on insights into the butterfly's flapping mechanism, we design a simplified thoracic structure shown in Fig.\ref{fig:3d}-(b). Biological muscles generate tension only through contraction and cannot actively produce thrust during relaxation, thus necessitating two pairs of muscles for thoracic contraction and relaxation. 
To streamline this mechanism, our design employs a crank-slider driven by twisted rubber bands, enabling the thoracic base to move back and forth, effectively simplifying the structure by reducing redundancy.

We utilize rigid polylactic acid (PLA) for the structural skeleton and flexible thermoplastic urethane (TPU) for the hinges, fabricating the integrated structure via FDM 3D printing technology. 
Initially, the wings remain horizontal with no deformation in the TPU hinges. As the thoracic base moves upward, the TPU hinges experience varying degrees of deflection, driving the wings upward and storing elastic energy that increases rapidly with further deformation. 
The stored elastic energy reaches a maximum at the wings' highest position and is subsequently released as the wings return to the neutral position during the downward stroke. The TPU hinges similarly function during downward movement. 

\subsection{Abdominal Undulation Mechanism}
In butterfly takeoff flight, the abdomen swings upward during the wing downstroke, and downward during the wing upstroke, with both motions occurring at the same frequency but in antiphase \cite{Fujikawa-2008}.

Based on this characteristic, we design a rotatable TPU pivot positioned at a specific distance from the thorax. A carbon fiber rod (0.8 mm diameter) passes through this pivot, connecting at one end to the moving thoracic base via a flexible TPU hinge capable of contraction and bending, with the other end remaining free to swing shown in Fig.\ref{fig:3d}-(c). The TPU hinge contraction compensates for the varying distance between the thoracic base and the pivot during movement, while its bending capability mitigates tangential stress on the carbon fiber rod caused by the thoracic base's vertical displacement. At the rod's free end, PLA hollow rings weighing 0.4 g each are arranged at intervals to emulate the shape and mass of a butterfly's abdomen. Leveraging this setup, when the thoracic base moves downward for the wing downstroke, the abdomen swings upward, and vice versa, creating a coupled wing-abdomen motion mechanism.

\begin{figure*}[h]
    \centering
    \includegraphics[width=0.9\textwidth]{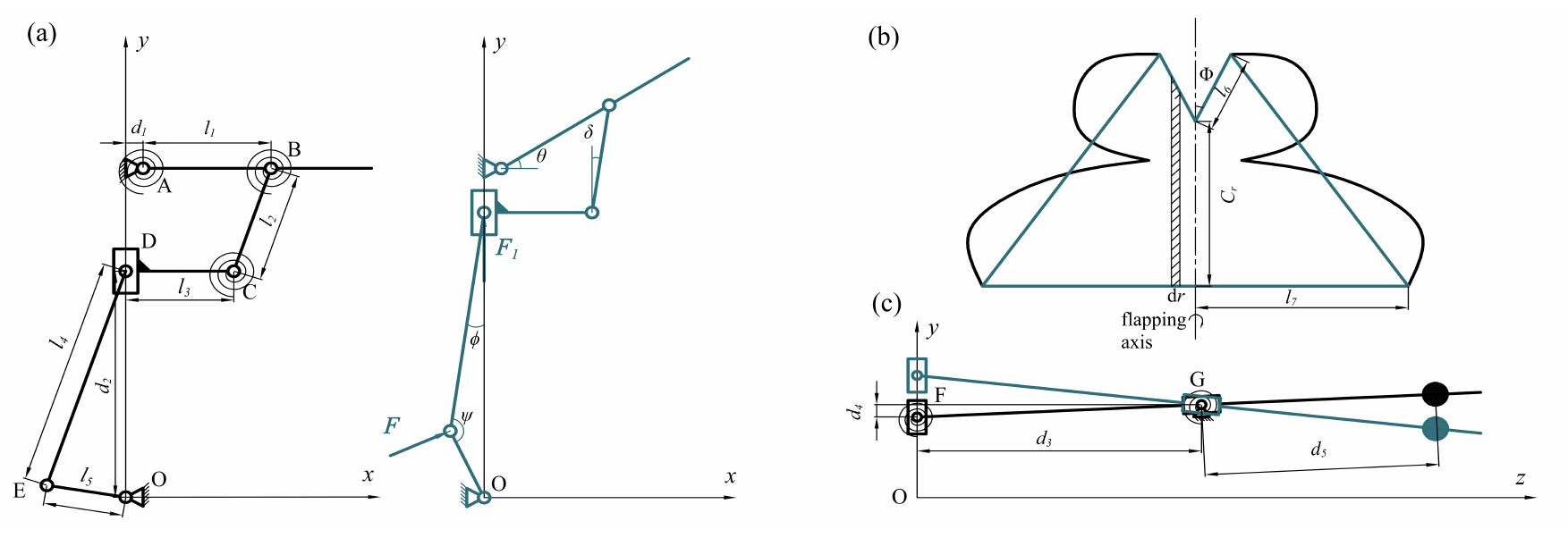} 
    \caption{(a) Pseudo-rigid model of the half-thorax in static equilibrium (black) and activated state (green). (b) Schematic drawing depicting the geometric parameters of the wing model and simplified outline (green). (c) Simplified model of abdomen mechanism in static state (black) and upstroke state (green).}
    \label{fig:mechanism}
\end{figure*}

\subsection{Wing Design}
Butterfly wings exhibit significant multidirectional flexible deformation during flight. Through observational and modeling experiments, Senda found that upward convex deformation during the downstroke generates increased lift, while downward convex deformation during the upstroke reduces drag, thereby enhancing aerodynamic performance \cite{Senda-2012-1}. Additionally, passive torsion of wing induced by aerodynamic forces contributes to improved flight stability.

Based on the characteristics described above, we design butterfly-inspired wings with integrated forewings and hindwings, using Expanded Polyethylene Foam (EPE) film as the membrane material shown in Fig.\ref{fig:3d}-(d). The leading edge of each wing is connected to a 1.0 mm diameter carbon fiber rod attached to a PLA wing base, providing spanwise flexibility during flapping. The inner edge of the hindwing is linked to a fixed 0.8 mm diameter carbon fiber rod on the frame, enabling wing torsion during the flapping motion.

\section{Dynamic Model Analysis}
\subsection{Flapping Mechanism Analysis}
Our flapping mechanism exhibits nonlinear behavior due to flexible hinge deformation and large linkage rotations; therefore, we employ a pseudo-rigid model for structural simplification. A symmetric half of the thorax is modeled as a pseudo-rigid structure comprising rigid linkages and torsional spring joints shown in Fig.\ref{fig:mechanism}-(a). In this model, the thorax base is represented by a slider on the symmetry plane connected via a linkage of length $l_3$, while the wing base and thoracic wall are represented by two linkages of lengths $l_1$ and $l_2$, respectively. The flexible TPU hinges between rigid plates are simplified as pin joints combined with torsional springs. Each equivalent linkage length includes the plate length plus half of the adjacent hinge lengths. To further simplify, we neglect the rigid portion of the half-flexible hinge connected to both the wing base and the thoracic base. All torsional springs connecting the wing base to the tergum ($\mathrm{A}$) , the wing base to the thoracic wall ($\mathrm{B}$), and the thoracic wall to the thoracic base ($\mathrm{C}$) share the same rotational stiffness, $K$.

When unloaded, the thoracic base is statically positioned at the neutral point $\mathrm{D}$, with the wing remaining horizontal. The angle ($\Theta$) between the thoracic wall linkage ($l_2$) and the x-axis is given by:
\begin{align}
\label{eq:Theta}
\Theta=\arccos\left(\frac{l_1+d_1-l_3}{l_2}\right)
\end{align}

The torque $\tau$ from the rotating rubber band generates a force $F=\tau/l_5$ at point $E$, perpendicular to the crank ($l_5$). The angle ($\psi$) between the crank ($l_5$) and the slide linkage ($l_4$) follows:
\begin{align}
\label{eq:cos_psi}
\cos \psi = \frac{l_4^{2}+l_5^{2}-(d_2+x)^{2}}{2l_4l_5}
\end{align}
The angle ($\phi$) between the slide linkage ($l_4$) and vertical direction is given by:
\begin{align}
\label{eq:sin_phi}
\sin \phi = \frac{l_5 \sin \psi}{d_2+x}
\end{align}

To produce an upward wing stroke, a vertical force $F_1$ pushes the thorax base slider (D):
\begin{align}
\label{eq:F}
F_1 = F \sin \psi \cos \phi
\end{align}
The joint travels a displacement $x$ from the initial height $d_2$:
\begin{align}
\label{eq:x}
x = l_2 \sin \Theta - l_2 \cos \delta + l_1 \sin \theta
\end{align}

As the thoracic base rises, the wing base linkage ($l_1$) rotates by angle $\theta_\mathrm{A}$ (relative to horizontal), and the spring $\mathrm{B}$ rotates by angle $\theta_\mathrm{B}$:
\begin{align}
\label{eq:alpha}
\theta_\mathrm{B} = \left|\Theta - \left(\frac{\pi}{2} - \theta_\mathrm{A} - \delta\right)\right|
\end{align}
The spring $\mathrm{C}$ rotates by angle $\theta_\mathrm{C}$:
\begin{align}
\label{eq:beta}
\theta_\mathrm{C} = \left|\Theta - \left(\frac{\pi}{2} - \delta\right)\right|
\end{align}
where $\delta=\arcsin\left[(l_1\cos\theta_\mathrm{A}-l_3+d_1)/l_2\right]$.

Moment equilibrium equations for the wing base linkage ($l_1$) during downstroke and upstroke are: 
\begin{align}
\label{eq:ME_2u}
&\frac{\boldsymbol{F_1} l_1 \sin\Bigl(\tfrac{\pi}{2}-\theta_\mathrm{A}-\delta\Bigr)}
      {2 \sin\Bigl(\tfrac{\pi}{2} - \delta\Bigr)}
  - K(\theta_\mathrm{A} + \theta_\mathrm{B} + \theta_\mathrm{C}) \notag\\
&\quad- \boldsymbol{M_{F_{\mathrm{drag}}}}(\dot{\theta}_\mathrm{A}) - m_1gR_\mathrm{C}\cos\theta_\mathrm{A}
 = I_\mathrm{wing} \ddot{\theta}_\mathrm{A}
\end{align}

Here, $M_{F_\mathrm{drag}}$ denotes the resisting moment exerted by aerodynamic drag on spring $\mathrm{A}$. When $F_1$ acts upward, both $F_1$ and  $M_{F_\mathrm{drag}}$  are positive. Conversely, when $F_1$ acts downward, both $F_1$ and $M_{F_\mathrm{drag}}$ are negative. $I_\mathrm{wing}$ is the moment inertia of half of the wing about the flapping axis, $m_1$ is the mass of half of the wing and $R_\mathrm{C}$ is the spanwise position of the wing, including contributions from the EPE membrane and carbon fiber rod, detailed in Table \ref{tbl:wing}. According to \cite{Howell-2013}, the flexural hinge stiffness $K$ is defined as 
\begin{align}
\label{eq:K}
K=\frac{\mathrm{d} M}{\mathrm{d} \theta}=\frac{EI}{H}
\end{align}
where $E$ is Young’s modulus, $I=bt^{3}/12$ is the area moment of inertia, and $H$, $b$, and $t$ are the hinge’s length, width, and thickness, respectively.

Table \ref{tbl:flapping} summarizes current design parameters chosen to
produce a static wing stroke of $50^\circ$ upward and $-37^\circ$ downward given $\pm10mm$ of thorax base
reciprocation. The design is not dynamically optimized because the
dynamic response in terms of wing stroke is unknown.
\begin{table}[h]
\centering
\caption{Parametric values for the pseudo-rigid model}
\label{tbl:flapping}
\begin{center}
\begin{tabular}{|>{\centering\arraybackslash}m{3cm}||
                  >{\centering\arraybackslash}m{1cm}||
                  >{\centering\arraybackslash}m{1cm}||
                  >{\centering\arraybackslash}m{1.5cm}|}
\hline
\textbf{Parameter} & \textbf{Symbol} & \textbf{Value} & \textbf{Unit}\\
\hline
Length of wing base linkage & $l_1$ & $14.95$ & $\mathrm{mm}$\\
\hline
Length of thorax wall linkage & $l_2$ & $13.49$ & $\mathrm{mm}$\\
\hline
Length of thorax base linkage & $l_3$ & $13.75$ & $\mathrm{mm}$\\
\hline
Length of slide linkage & $l_4$ & $30$ & $\mathrm{mm}$\\
\hline
Crank length & $l_5$ & $10$ & $\mathrm{mm}$\\
\hline
Tergum length & $d_1$ & $4.89$ & $\mathrm{mm}$\\
\hline
Initial distance between slide and O & $d_2$ & $30$ & $\mathrm{mm}$\\
\hline
Rotational stiffness of spring & $K$ & $1.57$ & $\mathrm{mN \cdot m/rad}$\\
\hline
\end{tabular}
\end{center}
\end{table}
\subsection{Aerodynamic Analysis}
We use quasi-steady model to calculate the resisting moment $M_{F_\mathrm{drag}}$ exerted by the
aerodynamic drag on the spring $\mathrm{A}$ and aerodynamic lift $F_\mathrm{lift}$. Before discussing our quasi-steady model development, it is important to first identify the key parameters that describe the wing geometry and flow
conditions.

Since the forewings and hindwings of the butterfly have the same motion pattern, we treat both wings as a whole for aerodynamic analysis. In our design, due to the irregular outline of the wing’s outer edge, the part of the wing outside the line connecting the trailing edge and the outermost tip in the spanwise direction is not under tension. Therefore, we approximate the wing as a regular quadrilateral for analysis, and its area is roughly similar to that of the effective part shown in Fig.\ref{fig:mechanism}-(b). Wing geometry is defined by the span ($l_6$),  root chord ($c_\mathrm{r}$), length of inner edge of hindwing ($l_7$) and the angle between it and flapping axis ($\Phi$) given in Table \ref{tbl:wing}, while
mean chord is given by equation \eqref{eq:c}:
\begin{align}
\label{eq:c}
\overline{c}=\frac{c_rl_7\sin\Phi}{2l_6}+\frac{c_r}{2}+\frac{l_7\cos\Phi}{2}
\end{align}
and the radius of second moment of area, $R_2$, which is given by equation \eqref{eq:R_2}:
\begin{align}
\label{eq:R_2}
R_2=\sqrt{\frac{1}{S}\int_{0}^{b}cr^{2}\mathrm{d}r}
\end{align}
where $S$ is the planform area of half of the wing and $c$ is the chord length at any radial position $r$. 
\begin{table}[h]
\centering
\caption{Parametric values for the wing geometry}
\label{tbl:wing}
\begin{center}
\begin{tabular}{|>{\centering\arraybackslash}m{3cm}||
                  >{\centering\arraybackslash}m{1cm}||
                  >{\centering\arraybackslash}m{1.5cm}||
                  >{\centering\arraybackslash}m{1cm}|}
\hline
\textbf{Parameter} & \textbf{Symbol} & \textbf{Value} & \textbf{Unit}\\
\hline
Wing span & $l_6$ & $190$ & $\mathrm{mm}$\\
\hline
Length of inner edge of hindwing & $l_7$ & $100$ & $\mathrm{mm}$\\
\hline
Root chord length & $c_\mathrm{r}$ & $65$ & $\mathrm{mm}$\\
\hline
Angle between inner edge of hindwing and flapping axis & $\Phi$ & $10$ & $\mathrm{^\circ}$\\
\hline
Wing moment of inertia & $I_\mathrm{wing}$ & $3.98\times 10^{-6}$ & $\mathrm{kg\cdot m^2}$\\
\hline
Spanwise position of the wing &$ R_\mathrm{C}$ & $81.7$ & $\mathrm{mm}$\\
\hline
Mass of the wing & $m_1$ & $0.425$ & $\mathrm{g}$\\
\hline
\end{tabular}
\end{center}
\end{table}

Since our quasi-steady model is intended to
accommodate a broad spectrum of flow conditions, it is
important to identify the critical dimensionless numbers that affect force generation. 
Following the definition in \cite{Lee-2016}, the Reynolds number $Re$ is given by
\begin{align}
\label{eq:Re}
Re=\frac{\overline{c}U_\mathrm{ref}}{\nu}
\end{align}
where $\nu$ is fluid viscosity and $U_\mathrm{ref}$ is the reference velocity. Here, $U_\mathrm{ref}=fl$ where $f$ is the flapping frequency and $l$ is the distance traveled by a point at $R_2$ per flapping cycle. 
We use blade elements method to dividing the wing
span into multiple wing strips in a spanwise manner. Based on the quasi-steady model \cite{Chin-2016}, the aforementioned forces interact synergistically, but for conciseness, we only consider the translational drag $F_\mathrm{drag}$, opposing the
wing velocity and the translational lift $F_\mathrm{lift}$, acting perprndicular to the wing velocity. They can be calculated on each wing strips as:
\begin{align}
\label{eq:F_drag}
\Delta F_\mathrm{drag}=\frac{1}{2}\rho c\Vert \boldsymbol{u} \Vert^{2}\mathrm{C_\mathrm{D_\mathrm{t}}}\Delta r
\end{align}
\begin{align}
\label{eq:F_lift}
\Delta F_\mathrm{lift}=\frac{1}{2}\rho c\Vert \boldsymbol{u} \Vert^{2}\mathrm{C_\mathrm{L_\mathrm{t}}}\Delta r
\end{align}
where $\rho$ is the air density, $c$ is the chord length at any radial position $r$, $\Vert \boldsymbol{u} \Vert=r\dot{\theta}$ is the wing-strip velocity, $\Delta r$ is the wing-strip width, $\mathrm{C_\mathrm{D_\mathrm{t}}}$ and $\mathrm{C_\mathrm{L_\mathrm{t}}}$ are the translational drag and lift coefficients, respectively. Therefore, the translational drag $F_\mathrm{drag}$ and the translational lift $F_\mathrm{lift}$ can be calculate as:
\begin{align}
\label{eq:F_drag2}
F_\mathrm{drag}=\sum \Delta F_\mathrm{drag}
\end{align}
\begin{align}
\label{eq:F_lift2}
F_\mathrm{lift}=\sum \Delta F_\mathrm{lift}
\end{align}
and $M_{F_\mathrm{drag}}$ and the average lift $\overline{F}_\mathrm{lift}$ of half of the wing over a cycle can be calculate as:
\begin{align}
\label{eq:M_{F_{drag}}}
M_{F_\mathrm{drag}}(\dot{\theta})=\sum\Delta F_\mathrm{drag}r
\end{align}
\begin{align}
\label{eq:avgF_lift}
\overline{F}_\mathrm{lift}=\frac{1}{T}\int_{0}^{T}\sum \Delta F_\mathrm{lift}\mathrm{d}t
\end{align}
where $T$ represents the time for one cycle.

And based on \cite{Lee-2016}, we can specify the relationship
between the coefficients and the angle of attack (AoA) $\alpha$:
\begin{align}
\label{eq:C_D_t}
\mathrm{C_\mathrm{D_\mathrm{t}}}=\mathrm{C_\mathrm{D_0}}+\mathrm{A_\mathrm{D}}[1-\cos(2\alpha)]
\end{align}
\begin{align}
\label{eq:C_L_t}
\mathrm{C_\mathrm{L_\mathrm{t}}}=\mathrm{A_\mathrm{L}}\sin(2\alpha)
\end{align}
where the coefficient $\mathrm{C_\mathrm{D_\mathrm{t}}}$ and $\mathrm{C_\mathrm{L_\mathrm{t}}}$ are both functions of
Reynolds number $Re$:
$$
\mathrm{A_\mathrm{D}}=1.873-3.14Re^{-0.369},
$$
$$
\mathrm{C_\mathrm{D_0}}=0.031+10.48Re^{-0.764},
$$
$$
\mathrm{A_\mathrm{L}}=1.966-3.94Re^{-0.429}.
$$
Here, the AoA $\alpha$ between the chord and incident velocity is $70^\circ$ based on observations from the flight video.
\begin{figure}[ht]
	\includegraphics[width=0.9\linewidth]{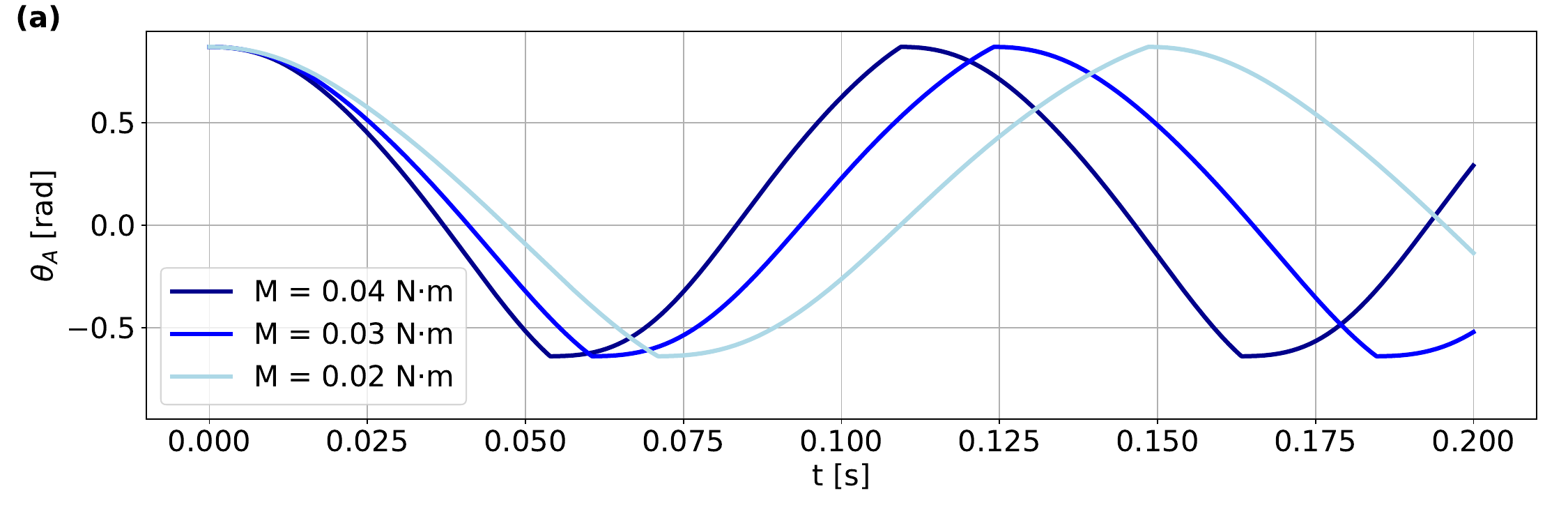}
	\includegraphics[width=0.9\linewidth]{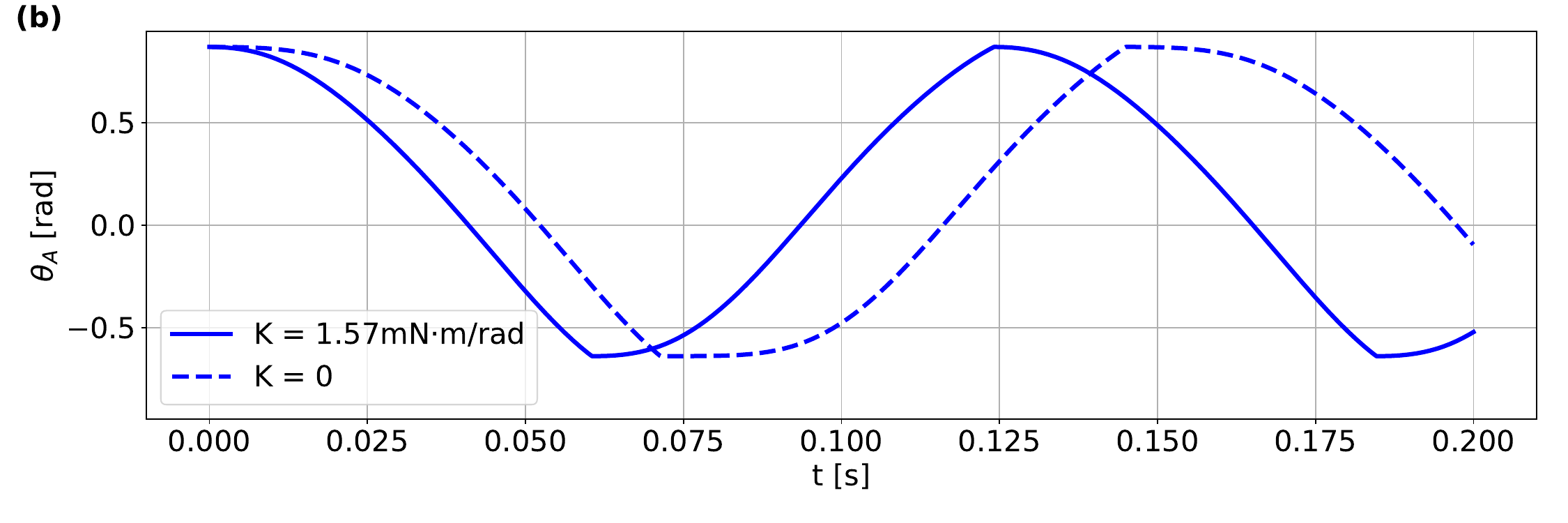}
     \centering
	\caption{(a) The time-dependent relationship curves of the wing base rotation angle $\theta_\mathrm{A}$ under different torque values. (b) The time-dependent relationship curves of the wing base rotation angle $\theta_\mathrm{A}$ with and without the use of flexible hinges.}
	\label{fig:theta}
\end{figure}

Based on the above method, we obtained the time-dependent relationship of the wing base rotation angle $\theta_\mathrm{A}$, under a given torque, which is illustrated in Fig.\ref{fig:theta}-(a). It was observed that as the applied torque decreases, the flapping frequency also decreases accordingly. When the rubber band was wound 80 times, generating an initial torque of approximately 0.03 $\mathrm{N\cdot m}$, flight experiments measured an initial flapping frequency of 8 Hz. This result closely matches our theoretical predictions, indicating the effectiveness of the simulation method. When the effect of the flexible hinges are ignored, the flapping frequency decreases significantly shown in Fig.\ref{fig:theta}-(b). Calculations show that the flexible hinges increase the flapping frequency by 16.85\% under otherwise identical conditions.
\subsection{Abdominal Undulation Mechanism Analysis}
\begin{figure}[h]
    \centering
    \includegraphics[width=0.9\linewidth]{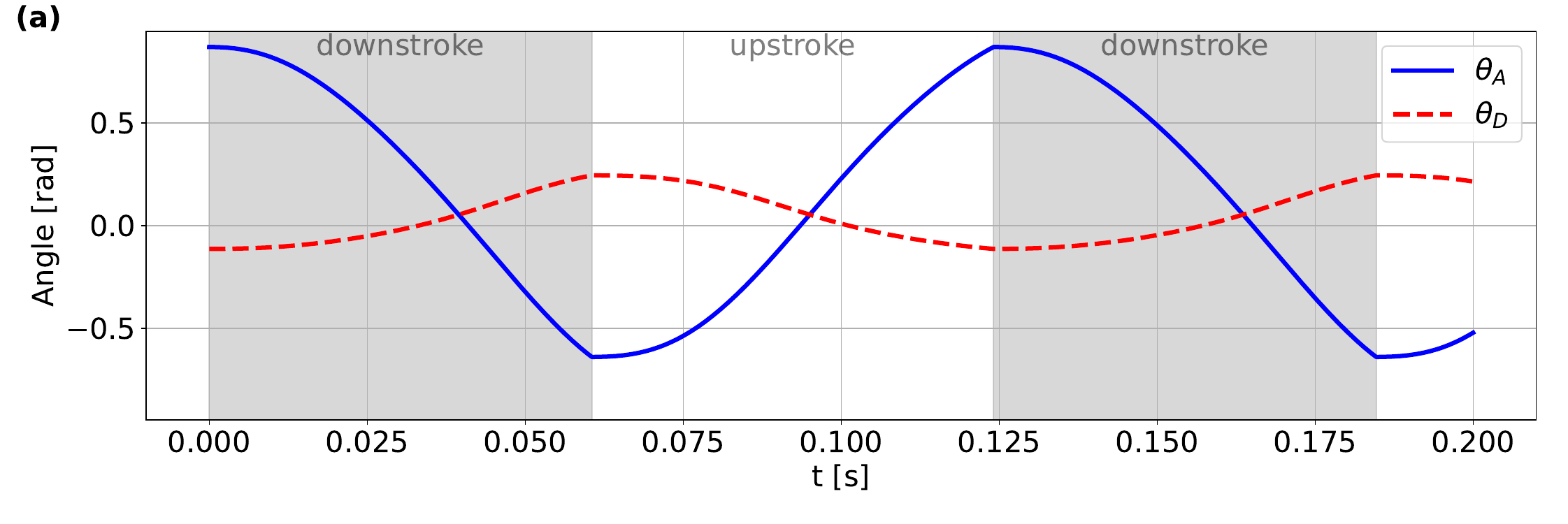}
    \includegraphics[width=0.9\linewidth]{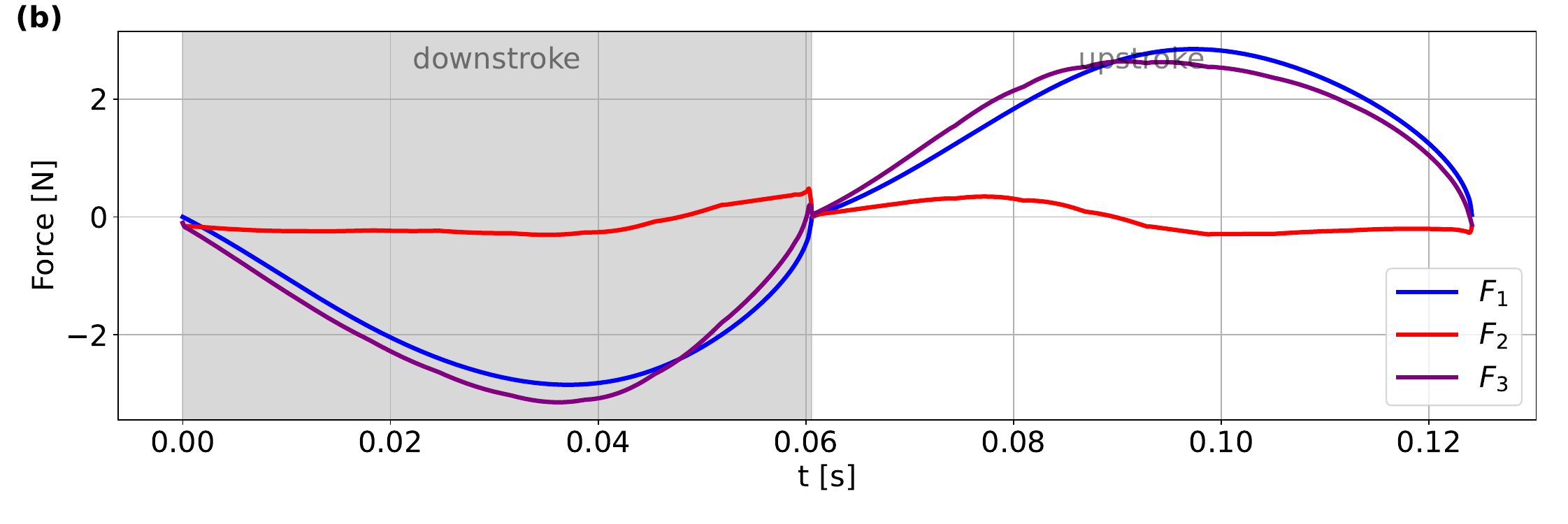}
    \includegraphics[width=0.9\linewidth]{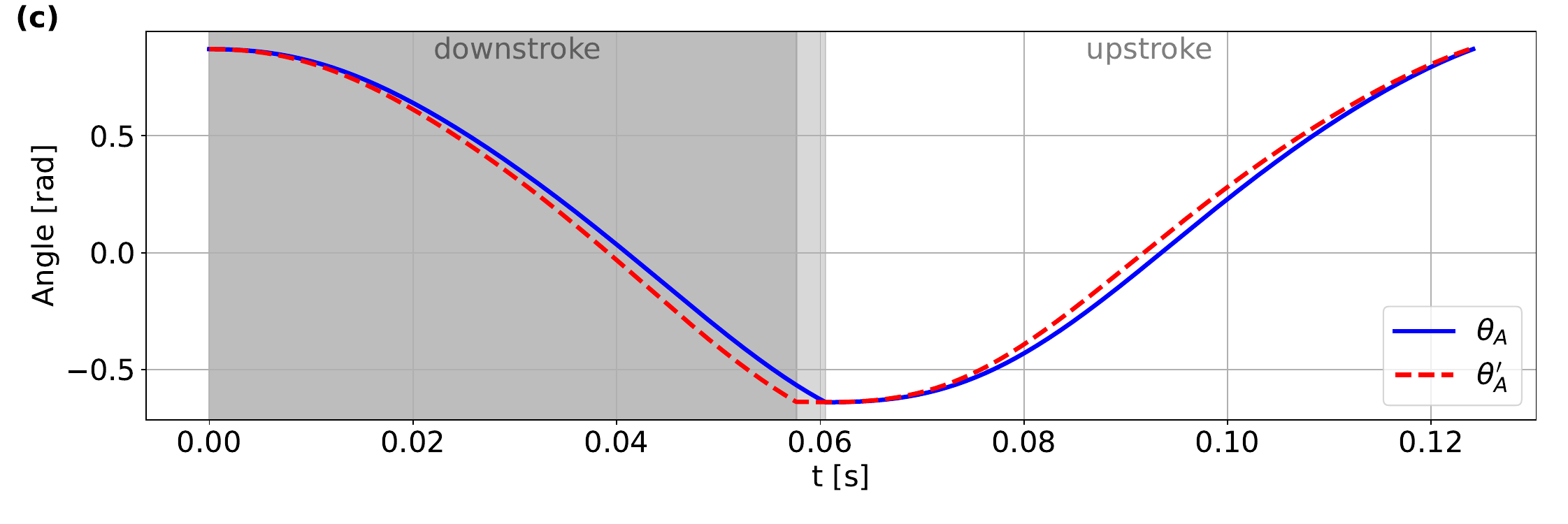}
    \includegraphics[width=0.9\linewidth]{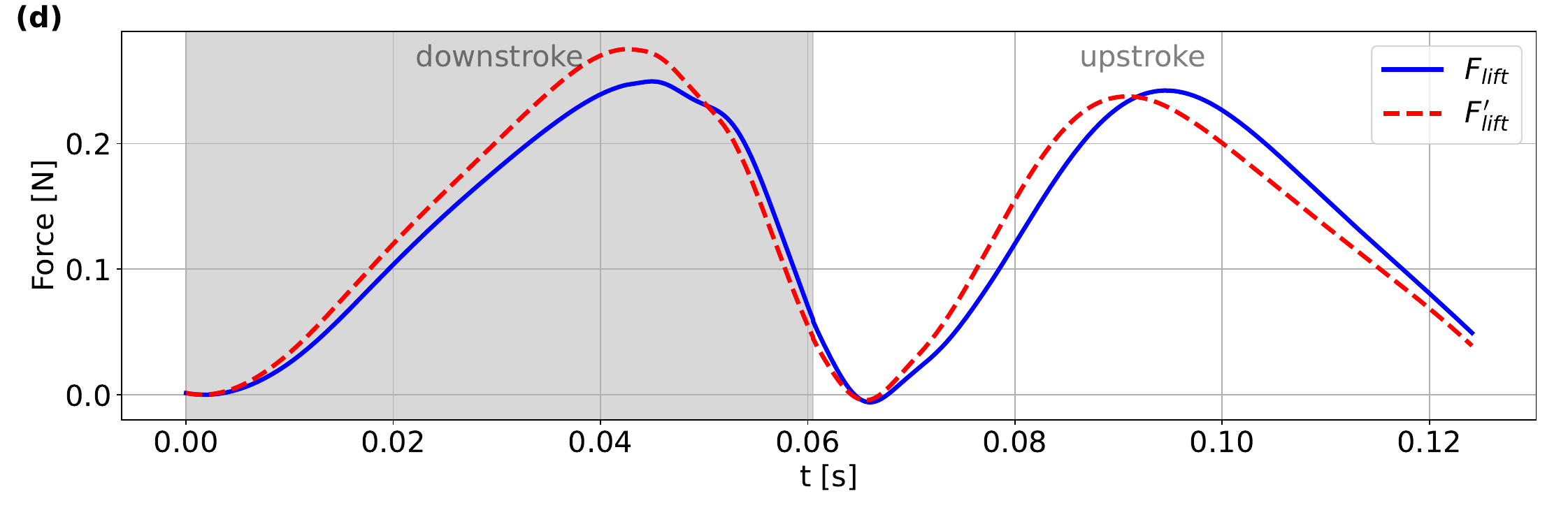}
    \caption{(a) The relationship between the abdominal rotation angle $\theta_\mathrm{D}$ and the wing base rotation angle $\theta_\mathrm{A}$. (b) The relationship between the vertical force $F_1$ exerted on the slider by a 0.03 $\mathrm{N\cdot m}$ torque and the vertical force $F_2$ generated by abdominal undulation, as well as their resultant force $F_3$. 
    (c) The time-dependent relationship of the wing base rotation angle $\theta_\mathrm{A}^\prime$ under the resultant force $F_3$.
    The discrepancy in alignment of the shaded regions for the two curves is attributed to the differing durations of the downstrokes.
    (d) The time-dependent relationship between the wing lift force with and without abdominal undulation, denoted as $F_\mathrm{lift}^\prime$ and $F_\mathrm{lift}$, respectively.}
    \label{fig:amended}
\end{figure}
We establish a model for abdomen undulation, which is shown Fig.~\ref{fig:mechanism}-(c). The flexible pivot is fixed to the fuselage, positioned \( d_3 \) horizontally and \( d_4 \) vertically from the slider’s initial position. The thoracic base, connected to a 0.8 mm carbon fiber rod via a flexible hinge, contracts and bends without deforming. The abdominal mass is approximated as a point mass at a distance \( d_5 \) from the pivot.
These dimensions are provided in Table \ref{tbl:abdomen}.

We can obtain the displacement-time relationship $x(t)$ of the slider from the previous calculations. Using this, we can determine the angle $\theta_\mathrm{D}$ between the abdominal carbon rod and the horizontal direction:
\begin{align}
\label{eq:theta_D}
\theta_\mathrm{D}=-\arctan \left(\frac{x-d_4}{d_3}\right)
\end{align}
and the dynamic equation of the mass $m_2$ is
\begin{align}
-\frac{F_2[(x-d_4)^2+d_3^2]}{d_3d_5}-m_2g\cos\theta_\mathrm{D}=m_2d_5\ddot{\theta}_\mathrm{D}
\end{align}
where $F_2$ is the force generated in the vertical direction of the slider due to abdominal undulation.
\begin{table}[h]
\centering
\caption{Parametric values for the abdomen mechanism}
\label{tbl:abdomen}
\begin{center}
\begin{tabular}{|>{\centering\arraybackslash}m{3cm}||
                  >{\centering\arraybackslash}m{1cm}||
                  >{\centering\arraybackslash}m{1.5cm}||
                  >{\centering\arraybackslash}m{1cm}|}
\hline
\textbf{Parameter} & \textbf{Symbol} & \textbf{Value} & \textbf{Unit}\\
\hline
Horizontal distance between pivot and slider& $d_3$ & $55$ & $\mathrm{mm}$\\
\hline
Vertical distance between pivot and initial position of the slider & $d_4$ & $3.75$ & $\mathrm{mm}$\\
\hline
Distance between the center of mass of abdomen and pivot & $d_5$ & $150$ & $\mathrm{mm}$\\
\hline
Distance from the CoM to the foremost point of BRB & $d_\mathrm{c}$ & $60$ & $\mathrm{mm}$\\
\hline
Mass of abdomen & $m_2$ & $2$ & $\mathrm{g}$\\
\hline
\end{tabular}
\end{center}
\end{table}

Building upon the previously outlined methodology, it is obtained that the relationship between the abdominal rotation angle $\theta_\mathrm{D}$ and the wing base rotation angle $\theta_\mathrm{A}$ shown in Fig.\ref{fig:amended}-(a). 
The results show that our model closely resembles the characteristics of real butterflies: when the wings flap downward, the abdomen swings upward, and when the wings flap upward, the abdomen swings downward.After obtaining the time-dependent curve of the abdominal rotation angle $\theta_\mathrm{D}$, we can derive the force required to sustain this motion. From this, we determine the reaction force $F_2$ acting in the vertical direction on the slider. By adding $F_2$ to the vertical force $F_1$ generated by the 0.03 $\mathrm{N\cdot m}$ torque on the slider, we obtain the corrected resultant force $F_3$ shown in Fig.\ref{fig:amended}-(b). By substituting $F_3$ into the previous aerodynamic force calculations for the wings, we obtain the corrected time-dependent curve of the wing base rotation angle $\theta_\mathrm{A}^\prime$. 
The results indicate that abdominal undulation has little effect on the flapping frequency but accelerates the downward wing stroke. Using Equations \eqref{eq:F_lift} and \eqref{eq:F_lift2}, we compute the time-dependent lift force curves $F_\mathrm{lift}$ and $F_\mathrm{lift}^\prime$ shown in Fig.\ref{fig:amended}-(d). The results show that, under otherwise identical conditions, the average lift with abdominal undulation is 3.4\% higher than without the undulation.

\begin{figure}[h]
	\centering
	\includegraphics[width=0.9\linewidth]{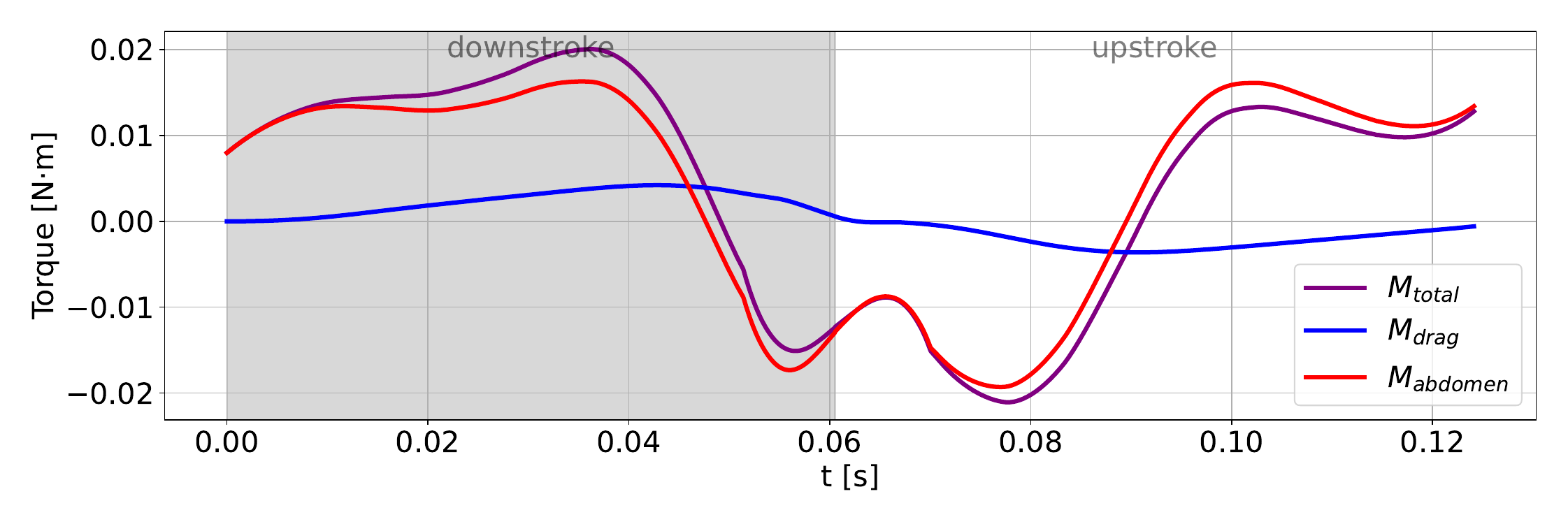}
	\caption{The time-dependent curves of the abdominal moment $M_\mathrm{abdomen}$, drag moment $M_\mathrm{drag}$, and their total moment $M_\mathrm{total}$.}
	\label{fig:moment}
\end{figure}

Subsequently, we analyze the moments generated by wing drag ($M_\mathrm{drag}$) and the inertial force of abdominal undulation ($M_\mathrm{abdomen}$) on the BRB’s center of mass (CoM) to predict the variation in pitch angle during flight. The moment is calculated according to the following equation:
\begin{align}
\label{eq:moment}
M_\mathrm{total}=F_\mathrm{drag}(d_\mathrm{c}-\frac{\overline{c}}{4})+m_2(g\cos\theta_\mathrm{D}+d_5\ddot{\theta}_\mathrm{D})d_5
\end{align}
where $\overline{c}/4$ represents the aerodynamic center of drag \cite{Sun-2014}, and $d_\mathrm{c}$ is the distance from the BRB’s CoM to its foremost point. The result is shown in Fig.\ref{fig:moment}.

The analysis reveals that wing drag moment on the BRB’s CoM fluctuates significantly due to abdominal inertia, amplifying pitch angle variation, consistent with experimental observations.

\section{Experiment Validation}
\subsection{Experiment Setup}
\begin{figure}[t]
   \centering
   \includegraphics[width=0.6\linewidth]{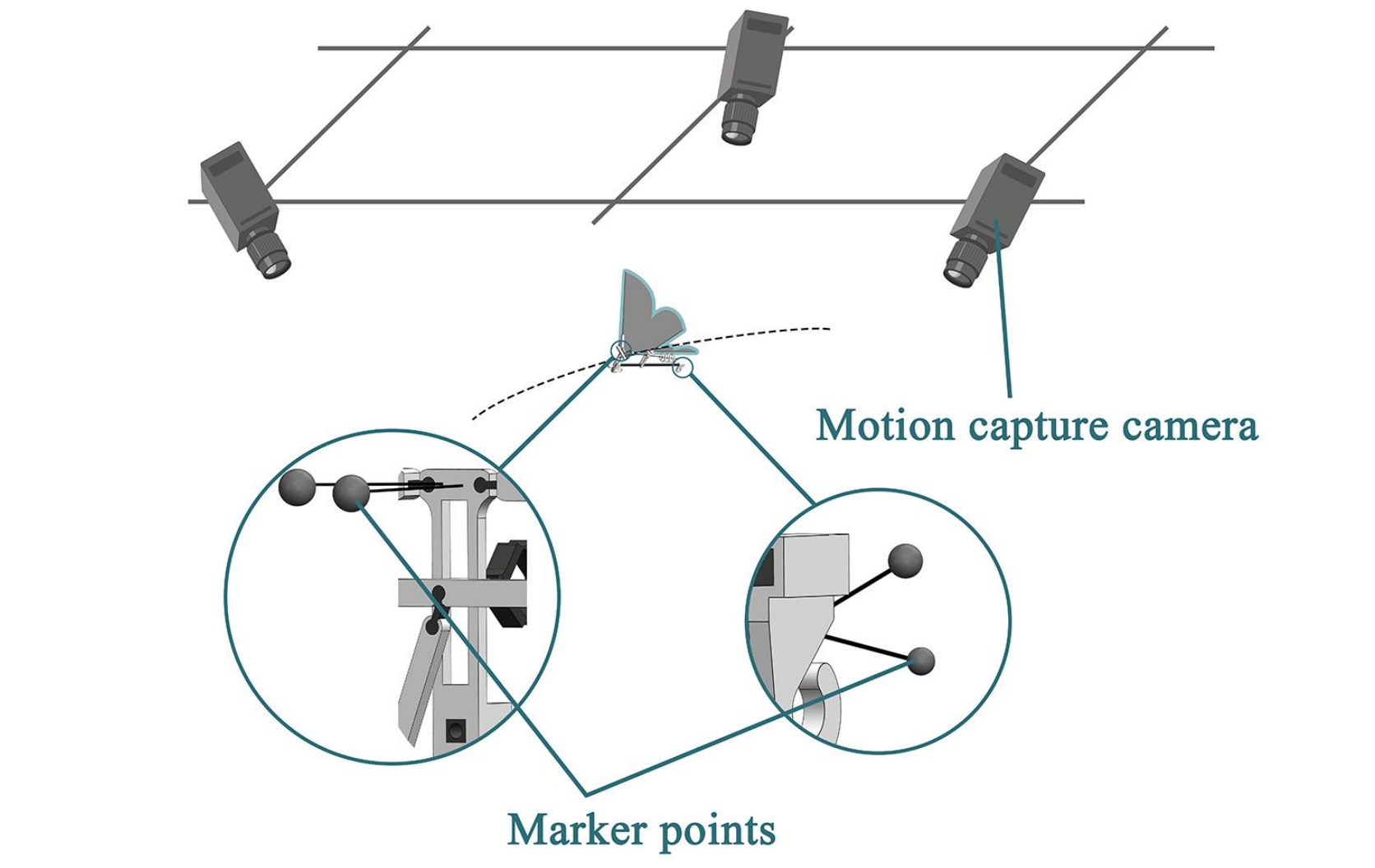}
   \caption{Schematic illustration of the BRB flight test under motion capture cameras. The setup includes multiple cameras tracking the model’s flight trajectory. Reflective marker points are attached to the tergum and tail hook at specific locations to facilitate motion capture and trajectory analysis.}
   \label{fig:makerponits}
\end{figure}

To verify the effect of abdominal undulation on the flight performance of BRB, we fabricate a prototype based on the aforementioned design\footnote{The mechanical drawings of the developed BRB are available at https://github.com/someNiKo/rubberbandBRB. And see the experiment video at https://youtu.be/68d5d4W\_XB4.}. The BRB has a wingspan of 19 cm, a root chord length of 6.5 cm, an inner hindwing edge of 10 cm, and a dihedral angle of $20^\circ$. The distance between the abdomen’s CoM and its pivot is 8 cm, with an abdominal weight of 1.8 g and a total weight of 12 g. The propulsion system consists of a 240 cm rubber band looped eight times, wound 80 turns to generate torque of 0.03 $\mathrm{N\cdot m}$. To facilitate motion capture, two reflective foam spheres (1 cm diameter) are mounted on the tergum at a horizontal separation, while another two identical spheres are attached to the tail hook at a vertical distance which is shown in Fig.\ref{fig:makerponits}. For comparative experiments, we fabricate two additional BRBs with identical parameters except for the abdominal configuration. One model lacks the PLA rings representing the abdomen, while in the other, the PLA rings are fixed at the end of a carbon fiber square rod, positioned at the same distance from the pivot as in the original design. This setup preserves the effect of the abdomen on the CoM of the BRB but eliminates its undulatory motion.

We conduct flight tests under motion capture cameras. Each BRB undergoes three flight trials. At the initial state, the wings are positioned at their maximum upstroke angle, and the fuselage is level with the ground. The BRBs are released from the same height with a slight initial velocity.

\begin{figure}[t]
    \centering
    \includegraphics[width=0.75\linewidth]{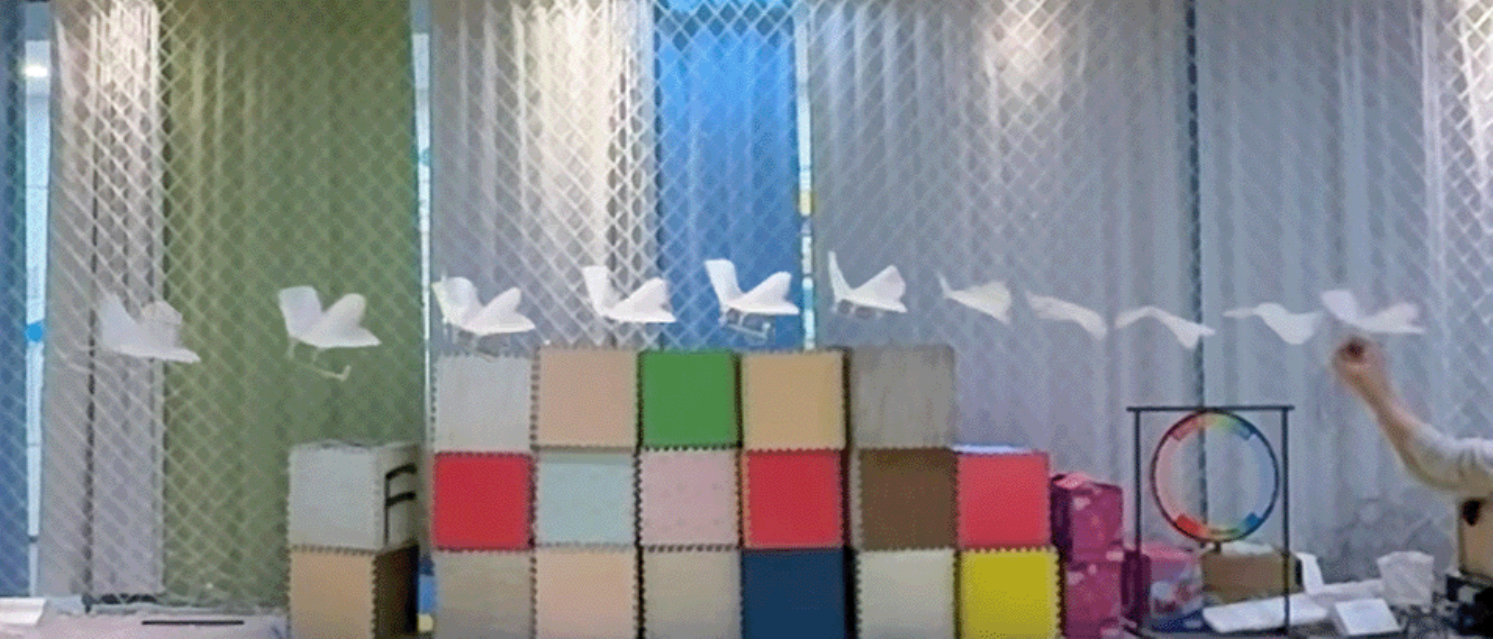}
    \caption{Real flight of the BRB in the motion capture arena.}
    \label{fig:tracjory}
\end{figure}
\subsection{Result \& Analysis}

Fig.\ref{fig:tracjory} presents one of the flight tests of the BRB with abdominal undulation. The BRB exhibits a relatively stable flight trajectory, maintaining altitude for a period of time and operates silently.

\begin{figure}[h]
    \centering
    \includegraphics[width=0.85\linewidth]{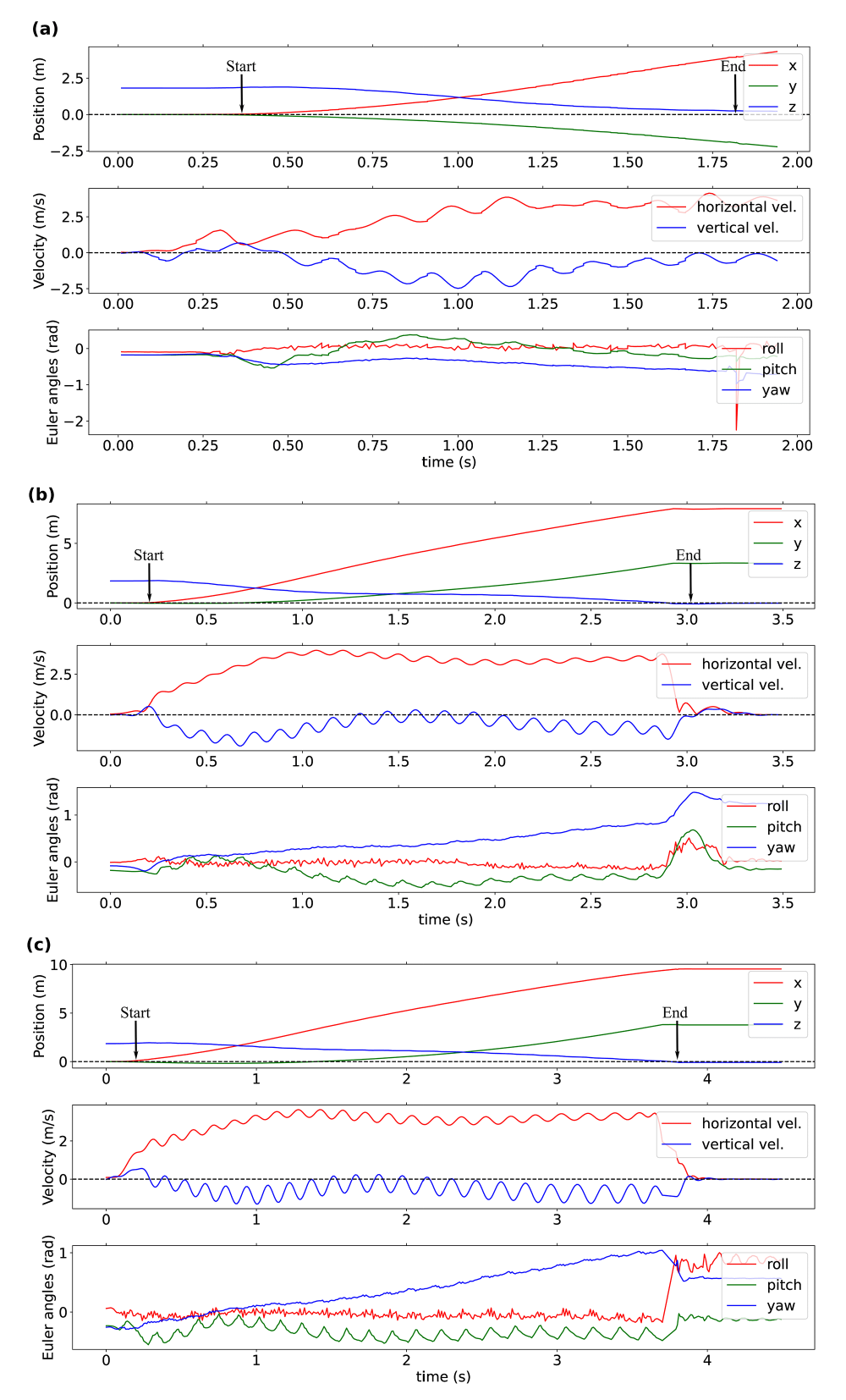}\\
    \caption{The position, velocity, and attitude data of the BRB without an abdomen (a), the BRB with abdominal load but without undulation (b), and the BRB with abdominal undulation (c) obtained using motion capture.}
    \label{fig:data}
\end{figure}

\begin{figure*}[h]
    \centering
    \includegraphics[width=0.9\textwidth]{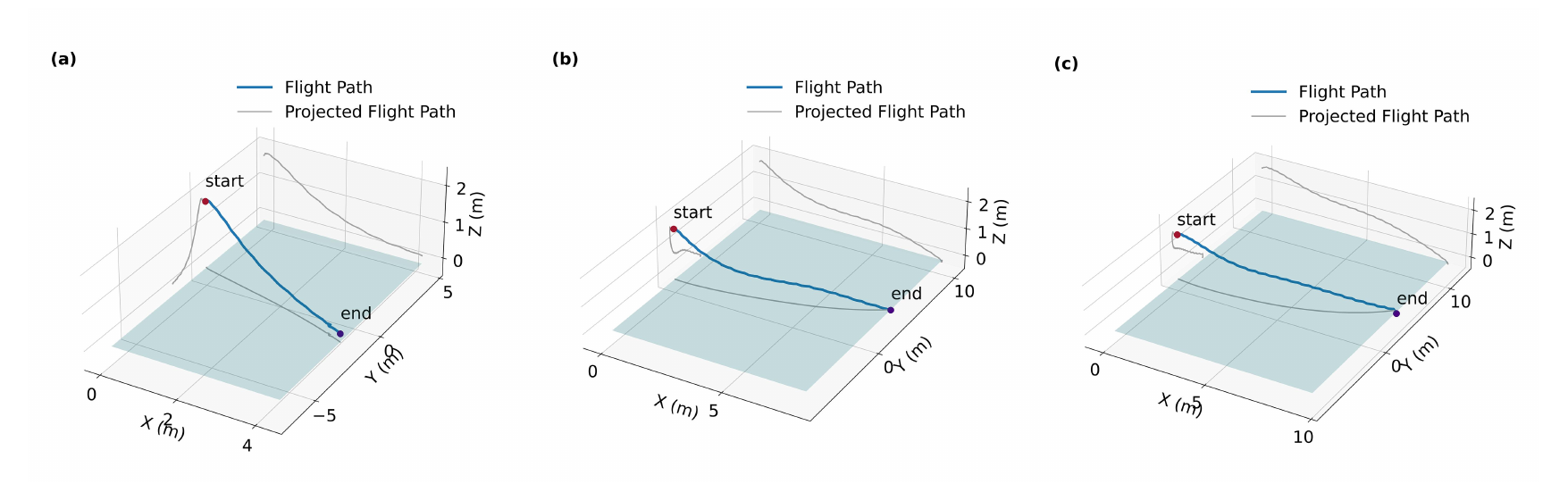} 
    \caption{The 3D view of the flight trajectory under motion capture, along with its projections on each coordinate plane of the BRB without an abdomen (a), the BRB with abdominal mass but without undulation (b), and the BRB with abdominal undulation (c).}
    \label{fig:3dline}
\end{figure*}

The motion capture data recorded for the three configurations are presented in Fig.\ref{fig:data}, and the 3D view of the flight trajectory are presented in Fig.\ref{fig:3dline}, where (a) represents the BRB without an abdomen, (b) represents the BRB with abdominal load but without undulation, and (c) represents the BRB with abdominal undulation. From the position data, it is evident that (c) achieved the greatest forward distance of 10 m, while (b) and (a) covered only 8 m and 5 m, respectively. Additionally, (c) was able to better maintain its flight altitude in the z-direction and sustained flight for approximately 4 seconds, whereas (a) only lasted 1 second and (b) remained airborne for 3 seconds. For vertical velocity, (c) has the slowest decay rate, and the velocity curve is smoother. The prolonged lift maintenance aligns with the theoretical prediction of a increase in average lift. 
For attitude analysis, we focus mainly on pitch angle. Both (c) and (b) exhibit significant oscillations, whereas (a) maintains a relatively smooth curves. This suggests that in (a), the wing lift fails to balance the BRB’s CoM, causing the body to tilt downward during flight. The oscillation amplitude of (c) is greater than that of (b), indicating that the inertial moment generated by abdominal undulation amplifies the oscillations, which aligns with the predictions of our theoretical model. 
The BRB with an undulating abdomen transitions more smoothly into stable flight after release, minimizing gravitational energy loss. 
During downstroke, the BRB tilts upward to capture more airflow and increase lift, while during upstroke it tilts downward to reduce drag. This oscillatory mechanism thus enhances flight efficiency.

\section{Conclusions}
This research presents a biomimetic robotic butterfly (BRB) incorporating a compliant mechanism to facilitate coupled wing-abdomen motion. Through theoretical modeling and empirical flight testing, it is demonstrated that the undulation of the abdomen contributes to increased lift production, stabilization of pitch oscillations, and enhanced overall flight performance. Motion capture analyses reveal that a BRB equipped with abdominal motion can achieve extended forward traversal, prolonged flight duration, and superior altitude retention. These results underscore the significance of abdominal movement in flapping-wing aerial vehicles (FWAVs) and offer valuable insights for developing future energy-efficient biomimetic flight designs.

\end{document}